\documentclass{article}

\usepackage{arxiv}
\usepackage{mathptmx}
\usepackage{cite}
\usepackage{amsmath}
\usepackage{times}
\usepackage{graphicx}
\usepackage{color}
\usepackage{multirow}
\usepackage{rotating}
\usepackage{bbm}
\usepackage{latexsym}

\usepackage{subfigure}
\usepackage{amssymb}
\usepackage{latexsym}
\usepackage{algorithm,setspace}
\usepackage{algorithmic}
\usepackage{bm}
\usepackage{mathrsfs}
\usepackage{color}

\usepackage{inputenc} 
\usepackage[T1]{fontenc}    
\usepackage{hyperref}       
\usepackage{url}            
\usepackage{booktabs}       
\usepackage{amsfonts}       
\usepackage{nicefrac}       
\usepackage{microtype}      
\usepackage{lipsum}

\title{A Novel Large-scale Ordinal Regression Model\thanks{This paper has been submitted to Neural Computing and Applications}}

\author{
  Yong Shi\\
  School of Economics and Management\\
  University of Chinese Academy of Sciences\\
  Beijing, China\\
  \texttt{yshi@ucas.ac.cn} \\
   \And
 Huadong Wang \\
  Research Center on Fictitious Economy $\&$ Data Science\\
  University of Chinese Academy of Sciences\\
  Beijing, China\\
  \texttt{wanghuadong14@mails.ucas.ac.cn} \\
 \And
 Xin Shen \\
  Department of Systems Engineering and Engineering Management\\
  The Chinese University of Hong Kong\\
  Hong Kong, China\\
  \texttt{xshen@se.cuhk.edu.hk} \\
  \And
 Lingfeng Niu\thanks{Corresponding author} \\
  School of Economics and Management\\
  University of Chinese Academy of Sciences\\
  Beijing, China\\
  \texttt{niulf@ucas.ac.cn} \\  
}

\begin{document}
\maketitle

\begin{abstract}
Ordinal regression (OR) is a special multiclass classification problem where an order relation exists among the labels. Recent years, people share their opinions and sentimental judgments conveniently with social networks and E-Commerce so that plentiful large-scale OR problems arise. However, few studies have focused on this kind of problems. Nonparallel Support Vector Ordinal Regression (NPSVOR) is a SVM-based OR model, which learns a hyperplane for each rank by solving a series of independent sub-optimization problems and then ensembles those learned hyperplanes to predict. The previous studies are focused on its nonlinear case and got a competitive testing performance, but its training is time consuming, particularly for large-scale data. In this paper, we consider NPSVOR's linear case and design an efficient training method based on the dual coordinate descent method (DCD). To utilize the order information among labels in prediction, a new prediction function is also proposed. Extensive contrast experiments on the text OR datasets indicate that the carefully implemented DCD is very suitable for training large data.
\end{abstract}


\section{Introduction}
Ordinal regression (OR) is a supervised learning problem that aims to learn a rule to predict labels of an ordinal scale, i.e., labels from a discrete but ordered set \cite{chu2007support}. In contrast to metric regression, it features finite ranks, among which the metric distances are not defined. Compared with ranks in classification, those in OR are different from the labels of multiple classes in classification problems due to the order information. Its prediction usually requires the predicted label as close as possible to the true label.
OR is generally considered as an intermediate problem between metric regression and multiclass classification.

 OR is widely employed in many domains\cite{evaluationofbreastcancer2005Cardosoa,fernandez2013addressing,wind,BCI,liu2009learning,Modeling2013Yoo,ColorImages2014Martin,CustomerChurn2008Rupesh,facialbeautyassessment2014HYan,facialexpressionrecognition2012Rudovic,Context-Sensitive2015Rudovic,ageestimation2011Chang}. In \cite{evaluationofbreastcancer2005Cardosoa}, a novel classification algorithm for ordinal data is adopted to objectively evaluate the aesthetic result of breast cancer conservative treatment. Fernandez-Navarro et al. \cite{fernandez2013addressing} addresses the sovereign rating problem by an ordinal regression approach as a result of its ordinal nature of dependent variable. In \cite{wind}, P.A. Gutiérrez et al. deem wind speed as a discrete variable and divides it into four levels, and thus it turns out to be a problem of classification.
 Yoon et al. \cite{BCI} propose an algorithm for adaptive, sequential classification in systems with unknown labeling errors for biomedical application of Brain Computer Interfacing. In facial recognition, Yan \cite{facialbeautyassessment2014HYan} proposes a cost-sensitive OR approach to predict the beauty information of face images captured in unconstrained environments which behaves like human beings; Rudovic et al. \cite{facialexpressionrecognition2012Rudovic} uses an ordinal manifold structure to encode temporal dynamics of facial expressions, ordinal relationships between their intensities and intrinsic topology of multidimensional continuous facial 
 data; Rudovic et al.\cite{Context-Sensitive2015Rudovic} developed Context-sensitive Conditional Ordinal Random Field model in estimating intensity levels
 of spontaneously displayed facial expressions; Chang et al. \cite{ageestimation2011Chang} uses relative order of age labels to  estimate age group and even precisely predict one's age only by facial images.
 All these works prove that utilizing the existing order information among labels in
modeling can significantly improve the prediction performance.

 \begin{figure}[h]
 \centering
  \includegraphics[width=3.5in]{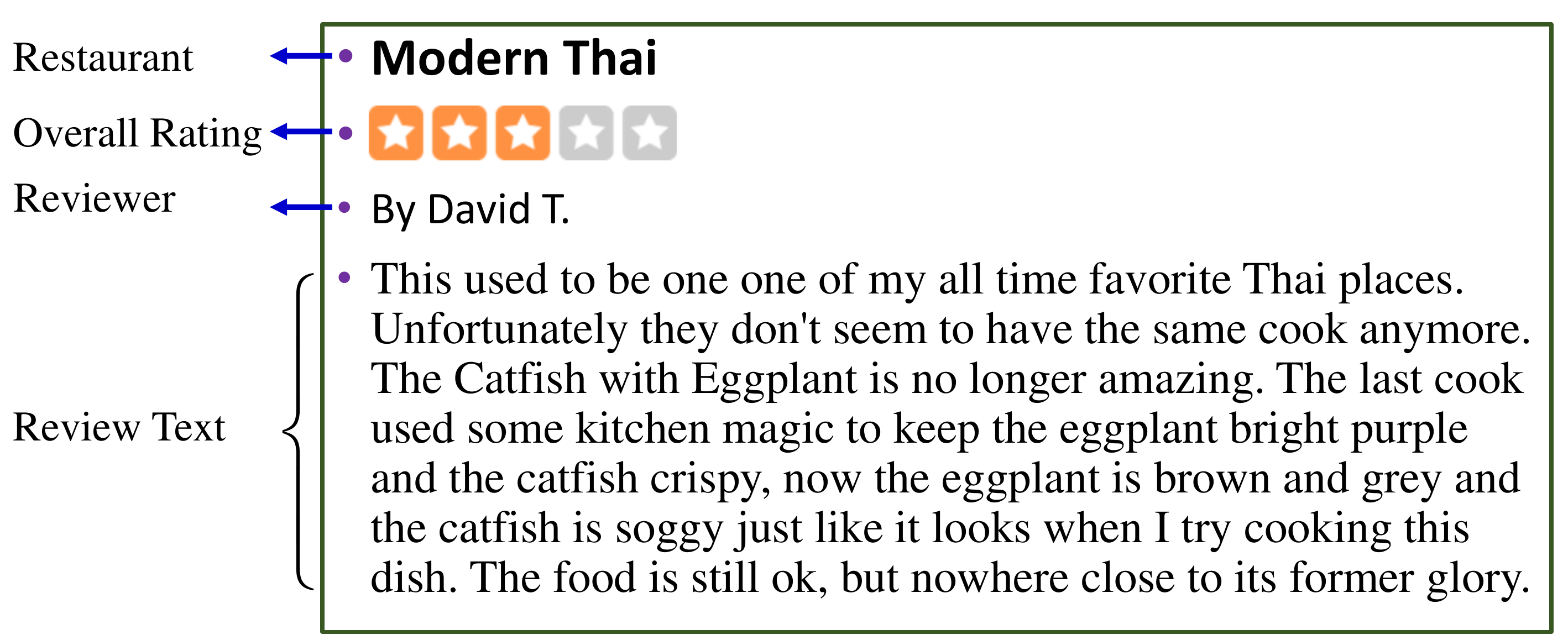}
  \caption{An example of restaurant reviews (from Yelp Dataset Challenge)}\label{fig:ORExample}
\end{figure}

 With the development of Internet and mobile communication technology, more and more people can freely express their opinions and personal preferences on all kinds of entities such as products and services. These reviews are useful for other users to make more sensible decisions, and for merchants to improve their service. A notable example is the restaurant's reviews which is shown in Figure \ref{fig:ORExample} (selected from Yelp Dataset Challenge\footnote{\url{https://www.yelp.com/dataset_challenge}}). The reviewer give review and corresponding rating about the restaurant by \{\emph{one star, two star, $\ldots$, five star}\}, depending on how much he or she likes it. The number of stars represent the different levels of preference. In Yelp dataset competition, one of the tasks is to predict the overall rating based on the review text, which can be used in the recommendation system. Obviously, this example can be treated as an OR problem. Under such circumstances, the 
 data of OR problem is usually in text type so that the feature vector is already high-dimensional which is extracted from text using n-gram words, such as word frequency vector or TF-IDF. To our knowledge, few researches focus on large-scale high dimensional OR problems, \cite{Baccianella2014Feature} has done some related works but it limited in feature selection. For OR problems, there still lacks relevant methods to be proposed and this paper aims to solve large-scale OR.

So far, a great deal of methods about OR have been proposed, such as SVOR \cite{chu2007support}, RedSVM \cite{lin2012reduction}, GPOR \cite{chu2005gaussian}, KDLOR \cite{sun2010kernel}, SVMOP \cite{waegeman2009ensemble}, AL \cite{NIPS2017_6659} etc. However, the characteristics of large scale, high dimension and high sparsity make traditional nonlinear model difficult to deal with. So we are considering some faster and more efficient algorithms for the large-scale and high dimensional problems. For OR problems, Support Vector Machines (SVMs) \cite{Vapnik,Largemargin,herbrich1999support} have shown promising results. In addition, linear SVMs \cite{Hsieh2008A, LinearSVMs} provide state-of-the-art prediction accuracy and is quite efficiently when handlling problems with a large number of features in the field of text mining. Because feature vectors are already linearly separable and the structure of the linear model is relatively simple. Such phenomena supports the hypothesis that linear SVM is a better alternative method to solve large scale OR.

According to the taxonomy \cite{Ordinal2016IEEE}, OR methods mainly fall into three categories, naive approach, ordinal binary decomposition approaches and threshold models. Ordinal binary decomposition approaches and threshold models fail to consider the data distribution of different ranks properly and might lead to unreasonable results. To better utilize the distribution information, the novel model called Nonparallel Support Vector Ordinal Regression (NPSVOR) was proposed in \cite{Wang2016Nonparallel}, and the alternating direction method of multipliers(ADMM) has been designed for its nonlinear dual model. Numerical experiments have shown that NPSVOR is superior to other SVM-based methods. Therefore, it is necessary to further study linear NPSVOR so that it can be applied to text mining.

This paper has the contributions as follows:
\begin{itemize}
  \item We have studied the large scale ordinal regression problems. Consider the linear NPSVOR, the coordinate descent (CD) with careful design is provided. Since optimizing fewer variables, the algorithm is faster than  directly extending the DCD algorithm of LIBLINEAR to linear NPSVOR.
  \item Considering the order information existing in labels, we proposed a new prediction function, which is better than the prediction function on the minimal distance principle. Compared with the latter one, the proposed prediction function can effectively reduce the generation of ambiguous decision regions and can give the lower prediction error.
  \item Fifty text ordinal regression datasets are provided or collected for testing and comparing. Those datasets are collected from different areas including sentiment analysis, film reviews, product reviews and health consultation service.
\end{itemize}
The organization of this paper begins with a brief review of NPSVOR and the coordinate descent method (CD) in Section \ref{relatedwork}. In Section \ref{NPSVOR}, we discuss the linear NPSVOR and propose an efficient implementation of coordinate descent methods. We conduct experiments in Section \ref{experiment} on some collected large scale ordinal regression datasets. Section \ref{conclusion} concludes this work.

\section{Related Works}\label{relatedwork}
\subsection{Nonparallel Support Vector Ordinal Regression}

Consider an OR problem with $n$ training examples $\mathcal{S} = \{({\bm x}_{i},y_i)\}_{i=1,\cdots,n}$,
where each example ${\bm x}_{i}$ drawn from a domain $\mathcal{X}\subseteq \Re^m$ and each ordinal label $y_i$ is an integer from a finite set of consecutive integers $\mathcal{Y}=\{1, 2, \ldots, p\}$.

In \cite{Wang2016Nonparallel}, a SVM-based model, called  Nonparallel Support Vector Ordinal Regression (NPSVOR),  was proposed for OR problem and got a superior performance among the existing state-of-art SVM-based models. This method firstly constructs three index sets for each rank $k \in \{1, 2, \ldots, p\}$,
\begin{equation}\label{triple_decompsition}
 \mathcal{L}_k =\{i|\forall i,y_i<k\},\mathcal{I}_k =\{i|\forall i,y_i=k\},\mathcal{R}_k =\{i|\forall i,y_i>k\}.
\end{equation}
where $y_i$ is the rank of the instance ${\bm x}_i$. Its purpose is to find a linear hyperplane
$f_k(\bm{x})=\bm{w}_k^\top\bm{x}+b_k$ for each rank $k\in \mathcal{Y}$ based on the triple decomposition (\ref{triple_decompsition}). The model is constructed as the following optimization problem:
\begin{subequations}\label{model:prim bk}
\begin{align}
\mathop {\min }\limits_{{{\bm{w}}_k},b_k,{{\bm{\xi }}_k^+},{{\bm{\xi }}_k^-},{{\bm{\eta }}_k} } & \frac{1}{2}\|{{\bm{w}}_k}\|_2^2 + {C_1}\sum_{i\in \mathcal{I}_k} ({{\xi _{ki}^+}} +\xi_{ki}^-)  + {C_2}\sum_{i\notin \mathcal{I}_k} {{{\eta _{ki}}} }\\
\rm{s.t.}\quad\quad
 & -\varepsilon  - {\xi _{ki}^-}\leq {{\bm{w}}_k^\top}{{\bm{x}}_{i}}+b_k \le \varepsilon  + {\xi _{ki}^+},i\in \mathcal{I}_k,\label{model:prim:2}\\
&{{\bm{w}}_k^\top} {{\bm{x}}_{i}}+b_k \le -1 + {\eta _{ki}},i\in \mathcal{L}_k,\label{model:prim:3}\\
&{{\bm{w}}_k^\top} {{\bm{x}}_{i}}+b_k \ge 1 - {\eta _{ki}},i\in \mathcal{R}_k,\label{model:prim:4}\\
&{\xi _{ki}^+},{\xi _{ki}^-}\geq 0, i\in \mathcal{I}_k,\label{model:prim:5}\\
&{\eta _{ki}}\geq 0, i\notin \mathcal{I}_k. \label{model:prim:6}
\end{align}
\end{subequations}
where $C_1, C_2>0$ are model parameters.

\begin{figure}[h]
  \centering
    \includegraphics[width=2in,height=2in]{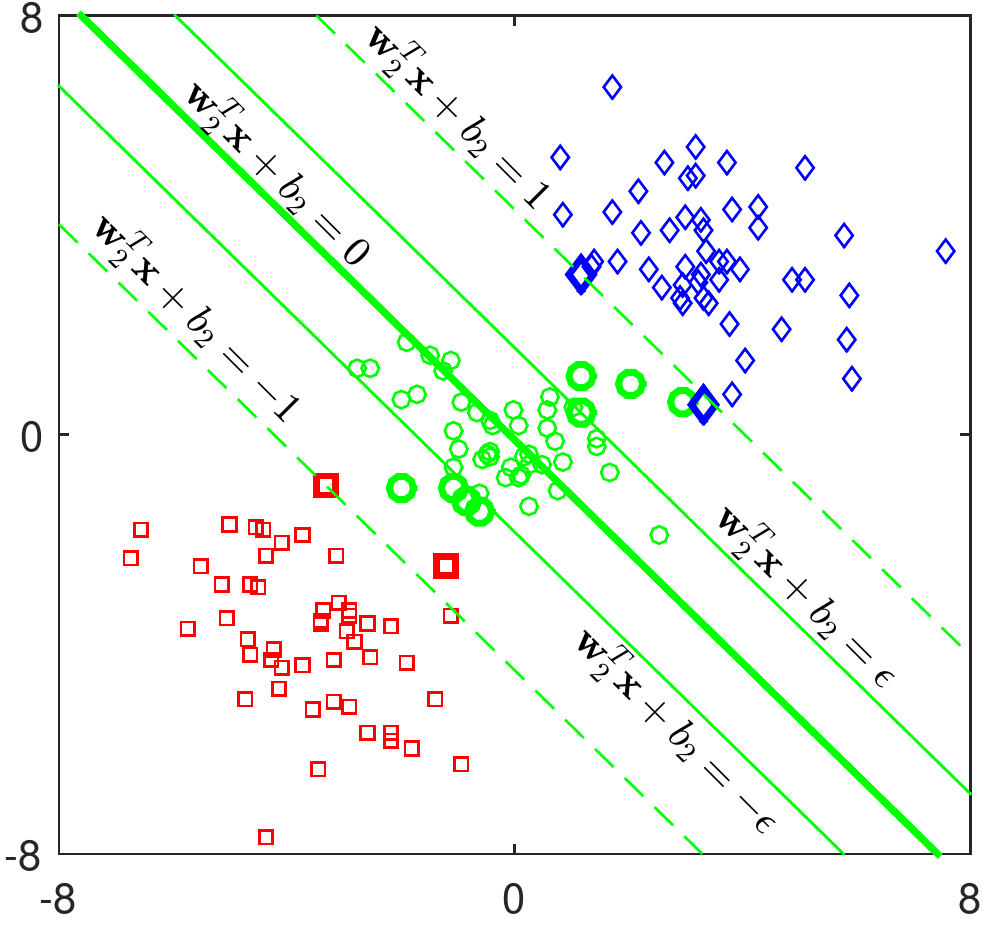}
\caption{The construction of proximal hyperplane for rank 2. ``\textcolor[rgb]{1.00,0.00,0.00}{$\square$}'',``\textcolor[rgb]{0.00,1.00,0.00}{ \scriptsize{$\bigcirc$}}'' and ``\textcolor[rgb]{0.00,0.00,1.00}{$\lozenge$}'' stand for rank 1,2,3, respectively. Support vectors are marked in bold.}
\label{fig:NPSVOR}
\end{figure}

Denote the optimal proximal hyperplane of rank $k$ obtained by (\ref{model:prim bk}) as $f_k(\bm{x})={\bm{x}}^\top {{\bm{w}}^*_k} + {b_k^*}=0$ for all $k=1, 2,\ldots, p$. In the model (\ref{model:prim bk}), the first term is a regularization item of $\bm{w}_k$, and both the second term and third term are Hinge losses. The second term requires learning the hyperplane $f_k(\bm{x})$ as close as possible to the $k$-th class samples $\{\bm{x}_i:i\in \mathcal{I}_k\}$. The third term requires that the samples with other ranks to be as far away as possible from the hyperplane $f_k(\bm{x})$. In order to utilize the order information contained among labels, the sample sets $\{\bm{x}_i:i\in \mathcal{R}_k\}$ and $\{\bm{x}_i:i\in \mathcal{L}_k\}$ are required to locate on both sides of the hyperplane.
Figure \ref{fig:NPSVOR} gives a geometrical illustration of the construction of proximal hyperplane in $\Re^2$.

Since NPSVOR is an extension model of the binary classification method Twin SVM \cite{Jayadeva2007Twin} or NPSVM \cite{tian2014nonparallel} in OR problems,  the prediction rule takes the minimal distance principle and is defined as
\begin{eqnarray}\label{nonlineardecisionfun}
r({\bm x}) = \arg {\min \limits_{k \in \{ 1, \ldots ,p\} }} |({{\bm{w}}^{*}_k})^\top{\bm{x}} + b^*_k|.
\end{eqnarray}
The OR predictor assigns the new sample $\bm{x}$ in the class which has the minimum absolute value of the learned function.

Wang et al. \cite{Wang2016Nonparallel} established the kernel NPSVOR model by introducing the kernel trick on its dual model, and designed the alternating direction method of multipliers (ADMM). But this algorithm only adapts to small-scale training data, and even with SMO, it is still difficult to handle large-scale  problem.

\subsection{Coordinate descent method}
Coordinate Descent Method (CD) is a common unconstrained optimization technique, which updates one variable at a time by minimizing a single-variable sub-problem. It is very popular recently for solving large-scale unconstrained optimization problems. Hong et al. \cite{Hong2017Iteration} provide a unified iteration complexity analysis for a family of general block coordinate descent methods and show that CD under block successive upper-bound minimization framework can achieve a global sublinear iteration complexity of $O(1/r)$, where $r$ is the iteration index.  CD has been exploited widely for linear SVM in large-scale scenarios. Hsieh et al. \cite{Hsieh2008A} proposed using CD for solving primal L2-SVM. Experiments show that their approach obtains an useful model  more quickly. Hsieh et al. \cite{Hsieh2008A} proposed a dual coordinate descent method (DCD) for linear SVM which investigated CDs for the dual problem of both SVM with hinge loss (L1-SVM) and SVM with the square hinge loss (L2-SVM). They proved that an $\varepsilon$-optimal solution can be obtained in $O(\log(1/\varepsilon))$ iterations. The shrinking strategy is also applied to make the CD faster.  Yuan et al. \cite{Yuan2010A} proposed a coordinate descent method for L1-regularized problems by extending Chang et al. \cite{chang2008coordinate}'s approach for L2-regularized classifiers. Tseng and Yun \cite{tseng2009coordinate} broadly discussed decomposition methods for L1-regularized problems. Ho and Lin \cite{ho2012large} extended LIBLINEAR's SVC solvers TRON and DCD to solve large-scale linear SVR problems. They are different because the former is to solve the primal problem, while the latter is for the dual problem.  In this paper, we aim at applying DCD to NPSVOR. A coordinate descent method updates one component of $\bm{\alpha}$ of its dual problem at a time by solving an one-variable sub-problem. It is competitive if one can exploit efficient ways to solve the sub-problem.

\section{Linear Nonparallel Support Vector Ordinal Regression}\label{NPSVOR}
In some applications, we include a bias term $b_k$ for each hyperplance in NPSVOR problems. To have a simpler derivation without the bias term $b_k$, one often augments each instance with an additional dimension:
\[{{\bm{x}}^\top} \leftarrow [{{\bm{x}}^\top},1],{{\bm{w}}_k^\top} \leftarrow [{{\bm{w}}_k^\top},b_k],k=1,\ldots,p.\]
The model is constructed as the following optimization problem:

\begin{align}\label{model:prim}
\mathop {\min }\limits_{{{\bm{w}}_k}} & \frac{1}{2}{\bm{w}}_k^\top{{\bm{w}}_k}+ {C_1}\sum\limits_{i \in {\mathcal{I}_k}}\max (|{\bm{w}}_k^\top{{\bm{x}}_i}| - \varepsilon ,0)\nonumber \\
 &+ {C_2}\sum\limits_{i \notin {\mathcal{I}_k}}\max (1-\tilde{y}_{ki}{\bm{w}}_k^\top{{\bm{x}}_i},0)
\end{align}
where $C_1, C_2>0$ are the model parameters, $k=1, 2,\ldots, p$, and $\tilde{y}_{ki} =1$, if $ i\in \mathcal{R}_k$, otherwise, $\tilde{y}_{ki} =-1$. It is obvious that the objective function is not differentiable.

 The optimal proximal hyperplane of rank $k$ obtained by (\ref{model:prim}) is denoted as $(\bm{w}^*_k)^\top\bm{x} = 0$.  The prediction rule can be written as follows
\begin{eqnarray}\label{decisionfun}
r({\bm x}) = \arg {\min \limits_{k \in \{ 1, \ldots ,p\} }} |(\bm{w}^*_k)^\top\bm{x}|.
\end{eqnarray}
This prediction function assigns the label based on the minimal distance from ${\bm{x}}$ to each nonparallel hyperplane.  But it does not consider the order information exists in ranks during prediction so that it easily leads to ambiguity in decision region. We will have a deeper discussion on it and propose a new predictor in Section \ref{Subsect:newPredictor}.

To emphasize that $p$ proximal hyperplanes are computed independently and are not necessarily parallel with each other, model (\ref{model:prim}) is named as Nonparallel Support Vector Ordinal Regression (NPSVOR) in this paper. Obviously, when the number of ranks is two, NPSVOR degenerates to nonparallel support vector machine for binary classification in \cite{tian2014nonparallel}.

From \cite{Wang2016Nonparallel}, the dual problem of (\ref{model:prim}) has the following form
\begin{equation} \label{dualmodel}
\begin{split}
\mathop {\min }\limits_{{{\bm {\alpha }}_k}} & {f_k}({{\bm {\alpha }}_k}) = \frac{1}{2}{\bm {w}}_k^\top {{\bm {w}}_k} + \varepsilon \sum\limits_{i \in {\mathcal{I}_k}} {(\alpha _k^ +  + \alpha _k^ - )}  - \sum\limits_{i \notin {\mathcal{I}_k}} {{\beta _{ki}}} \\
\text{s.t.} & 0 \le \alpha _{ki}^ + ,\alpha _{ki}^ -  \le {C_1},i \in {\mathcal{I}_k}\\
& 0 \le {\beta _{ki}} \le {C_2},i \notin {\mathcal{I}_k}
\end{split}
\end{equation}
where ${\bm{\alpha }}_k{\rm{ = }}{({({{\bm{\alpha }}_k^ - })^\top},{({{\bm{\alpha }}_k^ + })^\top},{{\bm{\beta }}_k^\top})^\top}$.

The primal-dual relationship indicates that primal optimal solution $\bm{w}_k$ and dual optimal solution $\bm{\alpha}_k^+$, $\bm{\alpha}_k^-$ and $\bm{\beta}_k$ satisfy
\begin{equation}\label{KTT_w}
{{\bm{w}}_k}=- \sum\limits_{i \in {\mathcal{I}_k}} {(\alpha _{ki}^ +  - \alpha _{ki}^ - ){{\bm{x}}_i}}  + \sum\limits_{i \notin {\mathcal{I}_k}} {{\tilde{y}_{ki}}{\beta _{ki}}{{\bm{x}}_i}}.
\end{equation}
An important property of the dual problem (\ref{dualmodel}) is that at the optimum condition,
\begin{equation}\label{alpha_property}
({\alpha}_{ki}^+)^*({\alpha}_{ki}^-)^*=0,\forall i\in \mathcal{I}_k.
\end{equation}
The dual problem of subproblem of NPSVOR for rank $k$ has $n+|\mathcal{I}_k|$ variables, while SVC has only $n$. With the increasing of the size of the training samples, the computational cost will increase significantly.  If a dual-based solver is applied without a careful design, the cost may be significantly higher than that of SVC.

\subsection{Dual Coordinate Descent Method for NPSVOR}\label{algorithm}
In this section, we introduce DCD, a coordinate descent method for the dual form of NPSVOR. We first extend the setting of \cite{Hsieh2008A} to NPSVOR and then propose a better algorithm using properties of NPSVOR.  Since each subproblem (\ref{dualmodel}) have same form, we disregard the subscript for rank $k$ for simplicity , such as.
some symbols ${{\bm{w}}_k},{\bm{\alpha }}_k^ + ,{\bm{\alpha }}_k^ - ,{{\bm{\beta }}_k},{\mathcal{I}_k}$ will be replaced by ${\bm{w}},{{\bm{\alpha }}^ + },{{\bm{\alpha }}^ - },{\bm{\beta }},\mathcal{I}$ in following discussion.

\subsubsection{A Direct Extension from the SVC's DCD solver to NPSVOR}
Let
${\bm{\alpha }}{\rm{ = }}{({({{\bm{\alpha }}^ - })^\top},{({{\bm{\alpha }}^ + })^\top},{{\bm{\beta }}^\top})^\top}$. Suppose the size of index set $\mathcal{I}$ be $l$,  the upper bound of the $i$-th variable of $\bm{\alpha}$ be
\[{U_i} = \left\{ {\begin{array}{ll}
{{C_1},}&{1 \le i \le 2l}\\
{{C_2},}&{2l< i \le n + l}
\end{array}} \right.\]
A coordinate descent method sequentially updates the $i$-th variable by solving the following subproblem.
\begin{equation}\label{D1:Subproblem}
\begin{split}
{\mathop {\min }\limits_d }\ &{{f}({\bm{\alpha }} + d{{\bm{e}}_i})}\\
{s.t.}\ &{0 \le {\alpha _i} + d \le {U_i}}
\end{split}
\end{equation}
 ${{\bm{e}}_i} \in {\Re^{n + {l}}}$  is a vector with $i$-th element one and others zero.

 We compute the
\begin{equation}\label{A_iB_i}
{A_i} = {\bm{x}}_i^\top{{\bm{x}}_i},{B_i} = {\bm{w}}^\top{{\bm{x}}_i},
\end{equation}
which is constants obtained using of the previous iteration $\bm{\alpha}$ .

 The objective function of (\ref{D1:Subproblem}) is a simple quadratic function of $d$:
\begin{equation*}
{f}({\bm{\alpha }} + d{{\bm{e}}_i}) = \frac{1}{2}\bar A_i{d^2} + G_id + \text{constants}
\end{equation*}
where
\begin{equation}\label{DCD-1:bar_Ai}
\bar A_i = \left\{ {\begin{array}{ll}
{{A_{i}}}  &{{\rm{if~}}1 \le i \le {l}}\\
{{A_{i - {l}}}} &{\rm{if~}} l<i\leq n+l
\end{array}} \right.
\end{equation}
and
\begin{equation}\label{DCD-1:Gi}
G_i = \left\{ {\begin{array}{ll}
-B_i + \varepsilon &{{\rm{if~}}1 \le i \le {l}}\\
{B_{i - {l}} + \varepsilon }&{{\rm{if~}}{l}< i \le 2{l}}\\
{B_{i - {l}} - 1}&{{\rm{if~}}2{l}< i \le n + {l}}
\end{array}} \right.
\end{equation}
In fact, $G_i$ is the $i$-th component of the gradient $\nabla f$. It is obvious that the optimal value $d$ of (\ref{D1:Subproblem}) can be solved in a closed form, so that $\alpha_i$ is updated by
\begin{equation}\label{D1:updatelphai}
{\alpha _i} \leftarrow \min ( {\max( {{\alpha _i} - \frac{G_i}{\bar A_i},0} ),{U_i}} )
\end{equation}

If ${\bar \alpha _i}$ is the current value and $\alpha_i$ is the value after the updating  by (\ref{D1:updatelphai}),then vector $\bm{w}$ can be maintained by
\begin{equation}\label{D1:Update_w}
{\bm{w}} \leftarrow \left\{ {\begin{array}{*{20}{l}}
{{\bm{w}} - ({\alpha _i} -{{\bar \alpha }_i}){{\bm{x}}_i}}& 1\leq i\leq l\\
{{\bm{w}} + ({\alpha _i} -{{\bar \alpha }_i}){{\bm{x}}_i}}& \text{~otherwise~}
\end{array}} \right.
\end{equation}
Hsieh et al. \cite{Hsieh2008A} check the violation of the optimality condition of (\ref{DualModel}) for the stopping condition because $\bm{\alpha}$ is optimal if and only if  $v_i=0$, which is defined as
\begin{align}\label{D1:Projection}
v_i = \left\{ {\begin{array}{ll}
{\min (G_i,0)}&{{\text{if~}}{\alpha _i} =0}\\
{\max(G_i,0)}&{{\text{if~}}{\alpha _i} = {U_i}}\\
G_i&{{\text{if~}}0 < {\alpha _i} < {U_i}}
\end{array}} \right.
\end{align}
If $v_i=0$ , then (\ref{D1:updatelphai}) and (\ref{D1:Projection}) imply that $\alpha_i$  needs not be updated. We show the overall procedure in Algorithm \ref{alg:LinearNPSVOR_full}.
\begin{algorithm}
\caption{DCD-1: a dual coordinate descent method for (\ref{model:prim})}
\label{alg:LinearNPSVOR_full}
\begin{algorithmic}[1] 
\STATE Initial $\bm \alpha=\bm 0 \in {\Re^{n + {l}}}$  and the corresponding ${\bm{w}} = \bm 0$.
\STATE Compute $A_i = \bm{x}_i^\top\bm{x}_i, i=1,\ldots,n$.
\WHILE{$\bm \alpha$ is not optimal,}
    \FOR {$i = 1, \ldots, n + {l}$}
        \STATE Obtain $\bar A_i$ and $G_i$ by (\ref{DCD-1:bar_Ai}) and (\ref{DCD-1:Gi}).
        \IF {$v_i \ne 0$,}
            \STATE Update ${{\alpha} _i}$  by (\ref{D1:updatelphai}).
            \STATE Update $\bm{w}$ by (\ref{D1:Update_w}).
        \ENDIF
    \ENDFOR
\ENDWHILE
\end{algorithmic}
\end{algorithm}

Hsieh et al. \cite{Hsieh2008A} apply a shrinking strategy to make Dual Coordinate Descent Method (DCD) faster in LIBLINEAR's SVC solvers. By gradually removing some variables, smaller optimization problems are solved to save the training time. They remove the variables which are likely to be bounded at optimum.  Although this DCD for SVC can be extended directly to solve the large-scale linear NPSVOR problem (\ref{dualmodel}) as given in Algorithm \ref{alg:LinearNPSVOR_full}, the procedure does not take NPSVOR's special structure into account. For examle. the number of variables of its dual model is larger than the size of training samples $n$, which obviously affects the efficiency of the model when the sample size is too large. Note that $\alpha _{i}^ +$ and $\alpha _{i}^ -$ in (\ref{dualmodel}) (omit the index $k$) are very related. We can see that in the following situations some operations using DCD directly are redundant.
\begin{itemize}
  \item The optimal solution $\bm{\alpha}$ of (\ref{dualmodel}) satisfies $\alpha_i^+\alpha_i^-=0, \forall i\in \mathcal{I}$ (The the rank index $k$ is omitted). Since $\alpha _{i}^ +$ and $\alpha _{i}^ -$ cannot be nonzero at the same time\footnote{From the KKT conditions, we have $(\bm{w}^\top\bm{x}_i+\varepsilon)\alpha _{i}^ +=0$ and $ (\bm{w}^\top\bm{x}_i-\varepsilon)\alpha _{i}^ -=0$. If $\alpha _{i}^ +$ and $\alpha _{i}^ -$ are none-zero at optimum, there exists $\bm{w}$ such that  $\bm{w}^\top\bm{x}_i+\varepsilon=0, \bm{w}^\top\bm{x}_i-\varepsilon=0$, but it is impossible.}, one of $\alpha _{i}^ +$ and $\alpha _{i}^ -$ is nonzero at optimum (without loss of generality, assume $\alpha _{i}^ +=0$), then the another one ($\alpha _{i}^ -$) must be zero. Thus, optimization of the variable $\alpha _{i}^ -$ is not necessary. We do not need to compute the corresponding $B_i=\bm{w}^\top\bm{x}_i$ and update $\bm{w}$. Therefore, some operations are wasted. Although shrinking strategy can partially solve this problem, alternatively we can explicitly use the property (\ref{alpha_property}) in designing the CD algorithm.
  \item Some operations in calculating gradient are unnecessary because the stopping condition needs the largest component of the optimal condition violation (\ref{D1:Projection}) of all variables,
      Suppose $\alpha _{i}^->0$ and $\alpha _{i}^ +=0$, if the optimality condition at $\alpha _{i}^ +$
       is not satisfied yet, then we have
      \[{v_{i + l}} = {G_{i + l}} = \tilde{y}_i{{\bm{w}}^\top}{{\bm{x}}_i} + \varepsilon  < 0,\]
      that is,
      \[0 <  - {v_{i + l}} <  - \tilde{y}_i{{\bm{w}}^\top}{{\bm{x}}_i} + \varepsilon  = {G_i}\]
      And since $v_i=G_i$, we have $|v_i| > |v_{i+l}|$. This shows us a larger absolute violation of the optimality condition occurs at $\alpha _{i}^-$. Thus, when $\alpha _{i}^->0$ and $\alpha _{i}^ +=0$, checking the $G_i$ and $v_i$ of $\alpha _{i}^ +$ is not necessary.
\end{itemize}
In the following discussion, we propose a method to address these issues.

\subsubsection{A New Coordinate Descent Method by Solving $\bf{a}^+$ and $\bf{a}^-$ Together}\label{section:DCD}
It should be noted that an important property of the dual problem (\ref{dualmodel}) is that at the optimum, $\alpha _{i}^ + \alpha_{i}^ -  = 0,i \in {\mathcal{I}}$, which, together with the condition $\alpha _{i}^ + ,\alpha _{i}^ -  \ge 0$, imply that at the optimum
\[\alpha _{i}^ +  + \alpha _{i}^ -  = |\alpha _{i}^ +  - \alpha _i^ - |, i \in {\mathcal{I}}.\]
 Denote ${{\bm{\alpha }}} = {({\alpha _{1}}, \ldots ,{\alpha _{n}})^\top},$
where
\[{\alpha _{i}} = \left\{ \begin{array}{ll}
\alpha _{i}^ +  - \alpha _{i}^ - &i \in {\mathcal{I}},\\
{\beta _{i}} &i \notin {\mathcal{I}}.
\end{array} \right.\]
Then $\bm{w}$ defined by (\ref{KTT_w}) becomes
\begin{equation}\label{WeightVector}
{\bm{w}} =-\sum\limits_{i \in {\mathcal{I}}} {(\alpha _{i}^ +  - \alpha _{i}^ - ){{\bm{x}}_i}}  + \sum\limits_{i \notin {\mathcal{I}}} {{\tilde{y}_{i}}{\beta _{i}}{{\bm{x}}_i}}= \sum\limits_{i = 1}^n {{\tilde y}_i}{{\alpha _i}{{\bm{x}}_i}}
\end{equation}
The problem (\ref{dualmodel}) can be transformed as:
\begin{equation}\label{DualModel}
\begin{split}
\min_{{\bm \alpha}}&f(\bm{\alpha}) = \frac{1}{2}{\bm{w }}^\top{{\bm{w}}} + \sum\limits_{i \in {\mathcal{I}}} {\varepsilon |{\alpha _{i}}|}  - \sum\limits_{i \notin {\mathcal{I}}} {{\alpha _{i}}} \\
s.t. &- {C_1} \le {\alpha _{i}} \le {C_1},i \in {\mathcal{I}}\\
&0 \le {\alpha _{i}} \le {C_2},i \notin {\mathcal{I}}
\end{split}
\end{equation}
where ${{\bm{\alpha }}} = {({\alpha _{i}}, \cdots ,{\alpha _{n}})^\top}$, the ${\bm{w}}$ get defined via ${{\bm{\alpha }}}$ as (\ref{WeightVector}).

A coordinate descent method sequentially updates one variable when fixing other variables.  Assume $\bm \alpha$ is the current iterate and its $i$-th component, denoted as a scalar variable $s$, is being updated.

(1) \textbf{For} $i \notin \mathcal{I}$, since
\[f({\bm{\alpha }} + (s-\alpha_i){{\bm{e}}_i})=\frac{1}{2}{A_i}{(s-\alpha_i)^2} + (B_i-1)s+ \text{const.}\]
 where $A_i = \bm{x}_i^\top\bm{x}_i, B_i={\tilde y_i}{{\bm{w}}^\top}{{\bm{x}}_i}$, which are constants obtained using $\bm \alpha$ of the previous iteration. So we just need to solve the following one-variable sub-problem:
\begin{equation}\label{SubPoblem_inotI}
\begin{split}
 \mathop {\min }\limits_{s\in [0,C_2]} &~h(s)\equiv \frac{1}{2}{A_i}{(s-\alpha_i)^2} + (B_i-1)s,
 \end{split}
 \end{equation}
If  $A_i > 0$, obviously the solution is:
\begin{equation}\label{UpdateAlpha_i}
{\alpha _i} \leftarrow \min \left( {\max \left( {{\alpha _i} - h'(\alpha _i)/{A_i},0} \right),{C_2}} \right).
\end{equation}
We thus need to calculate $A_i$ and $ h'(\alpha _i)$. First,
\begin{equation}\label{Derivative_h}
 h'(\alpha _i)= B_i-1.
\end{equation}

One can easily see that (\ref{SubPoblem_inotI}) has an optimum at $s=\alpha_i$ (i.e., no need to update $\alpha_i$) if and only if
$v_i = 0$,
where $v_i = h'_P(\alpha_i)$ means the projected gradient
\begin{equation}\label{ProjectedGradient_inotI}
h'_P(\alpha_i) = \left\{ {\begin{array}{*{20}{l}}
{h'(\alpha_i)}&{\text{~if~}0 < {\alpha _i} < {C_2},}\\
{\min (0,h'(\alpha_i))}&{\text{~if~}{\alpha_i } = 0,}\\
{\max(0,h'(\alpha_i))}&{\text{~if~}{\alpha _i} = {C_2}.}
\end{array}} \right.
\end{equation}
If  $h'_P(\alpha_i)=0$, then (\ref{UpdateAlpha_i}) and (\ref{ProjectedGradient_inotI}) imply that $\alpha_i$ needs not be updated.  Otherwise, we must find the optimal solution of (\ref{SubPoblem_inotI}).

(2) \textbf{For} $i\in \mathcal{I}$,  we solve the following one-variable sub-problem:
\begin{equation}\label{SubPoblem_inI}
\begin{split}
 \mathop {\min }\limits_{s\in [-C_1,C_1]} &~g(s)\equiv \frac{1}{2}{A_i}{(s-\alpha_i)^2} + B_is + \varepsilon|s|,
 \end{split}
 \end{equation}
 where $A_i = \bm{x}_i^\top\bm{x}_i, B_i={\tilde y_i}{{\bm{w}}^\top}{{\bm{x}}_i}$, which are constants obtained using $\bm \alpha$ of the previous iteration.

 Although the objective function of (\ref{SubPoblem_inI}) is not differentiable, we can still easily compute a simple closed-form solution to this problem by using subdifferential calculus \cite{rockafellar2015convex}. In detail,
 the objective function of  the subproblem (\ref{SubPoblem_inI}) can be represented as
 \[g(s) = \varepsilon |s| + \frac{{{A_i}}}{2}{\left( {s - {\alpha _i} + \frac{B_i}{A_i}} \right)^2} + {\rm{const}}{\rm{.}}\]
Thus, (\ref{SubPoblem_inI}) can be reduced to ``soft-thresholding" in signal processing and has a closed-form minimum.
Let  its derivatives at $s>0$  and $s<0$  be the following, respectively
\begin{align*}
{g'_p}(s) = A_i(s-\alpha_i) +B_i + \varepsilon, & \text{~if~}  s>0\\
{g'_n}(s) = A_i(s-\alpha_i) +B_i - \varepsilon, & \text{~if~} s<0
\end{align*}

If $\bm{x}_i\neq \bm{0}$ we have $A_i>0$, then both ${g'_p} (s) $and ${g'_n} (s)$ are strictly monotonic with respect to the variable $s$, i.e.
\[{g'_n}(s)<{g'_p}(s),\forall s\in \Re.\]
So $g(s)$ is a strictly convex quadratic function and has a unique minimum.

Since $g(s)$ is pairwise quadratic, we can consider the following three cases:
\begin{itemize}
\item If ${g'_p}(0)< 0$, the minimum $s^*$ of $g(s)$ will occur at range $s>0$. Let ${g'_p}(s)=0$, we have\\
    $s^*=\alpha_i-g'_p(\alpha_i)/A_i$;
\item If ${g'_n}(0)> 0$, the minimum $s^*$ of $g(s)$  will occur at range $s<0$. Let ${g'_n}(s)=0$, we have\\
    $s^*=\alpha_i-g'_n(\alpha_i)/A_i$;
\item If ${g'_n}(0)<0<{g'_p}(0)$, then the optimal solution of $g(s)$ can obtain at $s=0$.
\end{itemize}

It is easy to known that (\ref{SubPoblem_inI}) has the following closed form solution.
\begin{equation}\label{UpdateAlpha_inotI}
 {\alpha _i} \leftarrow \min \left( {\max \left( {{\alpha _i} + d_i, - {C_1}} \right),{C_1}} \right)
\end{equation}
where
\begin{equation}\label{GradientDirection}
 d_i = \left\{ {\begin{array}{ll}
 -g'_p(\alpha_i)/A_i,&{{{g'_p}}(\alpha_i) < A_i{\alpha_i}}\\
 -g'_n(\alpha_i)/A_i,&{{{g'_n}}(\alpha_i) > A_i{\alpha _i}}\\
-\alpha _i,&{{\rm{otherwise}}}
\end{array}} \right.
\end{equation}
In (\ref{GradientDirection}), we simplify the form of solution by using the property
\begin{equation}\label{Gradient_GpGn}
{g'_p}(\alpha_i) = B_i + \varepsilon \text{~and~} {g'_n}(\alpha_i) = B_i - \varepsilon
\end{equation}

If ${\bar \alpha _i}$  is the current value and ${\alpha _i}$  is the value after the update, we can maintain $\bm w$ by
\begin{equation}\label{UpdateW}
{\bm{w}} \leftarrow {\bm{w}} + ({\alpha _i} - {\bar \alpha _i}){\tilde y_i}{{\bm{x}}_i}.
\end{equation}

For the stopping condition, as a beginning, we study how to measure the violation of the optimality condition of (\ref{DualModel}) during the optimization procedure. After considering all situations, we know that
\begin{center}
$\alpha_i$ is optimal for (\ref{SubPoblem_inI}) if and only if $v_i=0$,
\end{center}
where
\begin{equation}\label{violation_vi}
v_i = \left\{ {\begin{array}{*{20}{l}}
{{{g'_p}}(\alpha _i)}&{0 < {\alpha _i} < {C_1}}\\
{{{g_n'}}(\alpha _i)}&{ - {C_1} < {\alpha _i} < 0}\\
{\min (0,{{g_n'}}(\alpha _i))}&{{\alpha _i} =  - {C_1}}\\
{\max (0,{{g_p'}}(\alpha _i))}&{{\alpha _i} = {C_1}}\\
{\max(0,{{g_n'}}(\alpha _i)) - \min (0,{{g'_p}}(\alpha _i))}&{{\alpha _i} = 0}
\end{array}} \right.
\end{equation}

\textbf{Stopping Condition and Shrinking Strategy}
Based on the above discussion, we can derive their stopping condition and shrinking scheme. We follow \cite{Yuan2010A} to use a similar stopping condition.
\begin{equation}\label{StopCondition}
{\left\| {{{\bm{v}}^t}} \right\|_1} < {\varepsilon _s}{\left\| {{{\bm{v}}^0}} \right\|_1}.
\end{equation}
where ${{\bm{v}}^0}$  and ${{\bm{v}}^t}$  are the initial violation and the violation in the $t$-th iteration, respectively. Note that  ${{\bm{v}}^t}$'s components are sequentially obtained via (\ref{violation_vi}) in $n$ coordinate descent steps of the $t$-th iteration.

For shrinking, we remove bounded variables (i.e.${\alpha _i} = 0,{C_2}, \forall i\notin \mathcal{I} $ or ${\alpha _i} = 0,{C_1}{\rm{~or~}}  - {C_1},\forall i\in \mathcal{I}$) if they may not be changed at the final iterations. Following \cite{Yuan2010A}, we use a ``tighter" form of the optimality condition to conjecture that a variable may have stuck at a bound.
We shrink $\alpha_i$ if it satisfies one of the following conditions:\\
For $i\notin \mathcal{I}$,
\begin{equation}\label{ShrinkingConditions_inotI}
  \alpha_i = 0  \text{~and~} v_i > M, \text{~or~} \alpha_i = C_2 \text{~and~} v_i <- M. \\
\end{equation}
and for $i\in \mathcal{I}$,
\begin{subequations}\label{ShrinkingConditions}
\begin{align}
&  {\alpha _i} = 0 \text{~and~}  {g'_n}({\alpha _i}) <  - M \text{~and~}  {g'_p}({\alpha _i})>M,\label{ShrinkingConditions:1}\\
& {\alpha _i} = C_1 \text{~and~}  {g'_p}({\alpha _i}) <  - M,\label{ShrinkingConditions:2} \\
&  {\alpha _i} = -C_1 \text{~and~}   {g'_n}({\alpha _i}) > M,\label{ShrinkingConditions:3}
\end{align}
\end{subequations}
where
\begin{equation}\label{Shrink_M}
M = \mathop {\max }\limits_{\forall i} |v_i^{t - 1}|
\end{equation}
is the maximal violation of the previous iteration.

Algorithm \ref{alg:LinearNPSVOR_detial} is the overall procedure to solve (\ref{DualModel}).
\begin{algorithm}
\setstretch{1.2}
\caption{DCD-2: DCD with a stopping condition and a shrinking implementation for rank $k$}
\label{alg:LinearNPSVOR_detial}
\begin{algorithmic}[1] 
\STATE Initial $\bm \alpha := \bm{0}$  and the corresponding ${\bm{w}} := \bm{0}$.
\STATE Compute $A_i = \bm{x}_i^\top\bm{x}_i, i=1,\ldots,n$.
\STATE $T \leftarrow \{ 1, \ldots ,n\} $, $M \leftarrow \infty$
\FOR{$t = 1,2, \ldots $}
    \STATE Randomly permute $T$.
    \FOR {${i\in T}$}
        \STATE Find ${\tilde y_i}$ by
        ${\tilde y_i} = \left\{ {\begin{array}{*{20}{l}}
            { - 1,}&{{y_i} \le k}\\
            {1,}&{{y_i} > k}
            \end{array}} \right.$
        \STATE $B={\tilde y_i}{{\bm{w}}^\top}{{\bm{x}}_i}$
        \IF {$y_i\neq k$,}
            \STATE $G = B - 1$
            \STATE $d_i \leftarrow -G/A_i$  and find $v_i^t$  by (\ref{ProjectedGradient_inotI})
            \IF {the condition in (\ref{ShrinkingConditions_inotI}) is satisfied,}
                \STATE $T \leftarrow T\backslash \{ i\}$, Continue
            \ELSE
                \STATE ${\bar \alpha _i} \leftarrow {\alpha _i}$, ${\alpha _i} \leftarrow \min (\max ({\alpha _i} + d_i, 0),C_2)$
            \ENDIF
        \ELSIF{$y_i= k$,}
            \STATE ${G_p} = B + \varepsilon $ and ${G_n} = B - \varepsilon$\\
            \STATE Find $d_i$  by (\ref{GradientDirection}) and find $v_i^t$  by (\ref{violation_vi})
            \IF {any condition in (\ref{ShrinkingConditions}) is satisfied,}
                \STATE  $T \leftarrow T\backslash \{ i\} $,  Continue
            \ELSE
                \STATE ${\bar \alpha _i} \leftarrow {\alpha _i}$,${\alpha _i} \leftarrow \min (\max ({\alpha _i} + d_i, -C_1),C_1)$
            \ENDIF
        \ENDIF
        \STATE ${\bm{w}} \leftarrow {\bm{w}} + ({\alpha _i} - {\bar \alpha _i}){\tilde y_i}{{\bm{x}}_i},$
        \IF {$\|{{\bm{v}}^t}\|_1 < {\varepsilon _s}\|{{\bm{v}}^0}\|_1$ and $|T|<n$}
        \STATE $T \leftarrow \{ 1, \ldots ,n\}$, $M \leftarrow \infty$
        \ENDIF
        \STATE $M \leftarrow  \mathop {\max }\limits_{\forall i} |v_i^{t}|$
    \ENDFOR
\ENDFOR
\end{algorithmic}
\end{algorithm}

\subsection{Constructing The Predictors Based on The Order Binary Classifications}\label{Subsect:newPredictor}
The NPSVOR model can be regarded as an extension of the TSVM model \cite{Jayadeva2007Twin}  which was constructed for binary classification problem, and many studies have shown that TSVM has better performance than the standard SVM on many binary classification problems. Its prediction function is defined as (\ref{decisionfun}) based on the minimum distance principle. For a given sample, it will be assigned to the rank where the nearest hyperplane is located. However, there are still several problems with this prediction criterion for ordinal regression:
\begin{enumerate}
  \item When these hyperplanes are not parallel, it is easy to produce ambiguous regions when predicting. At this situation, using this prediction function based on the minimum distance principle is unreasonable. It may assign the sample with a wrong label which have larger error cost.
  \item In prediction, the prediction function just takes the ordinal regression as a general multiple-class classification, and doesn't consider the order information exiting in labels.
\end{enumerate}

Lin and Li \cite{lin2012reduction} proposed a framework to deal with the ordinal regression, that is, the ordinal regression is decomposed into a series of ordered binary classifications, and then constructed a discriminant function for each of them. Here, each discriminant function can be seen as a preference function of binary judgment problem ``Is it better than the rank $k$?" and then based on these binary judgment to build rank prediction. Inspired by this idea, using the approximated hyperplanes  ${f_k}({\bm{x}}) = \bm{w}_k^\top{\bm{x}} (k = 1, \ldots ,p)$ of $p$ classes, we construct the following discriminant functions of $p-1$ binary classifiers
 \begin{equation}\label{NewHyperplane}
 {g_k}({\bm{x}}) = {f_k}({\bm{x}}) + {f_{k+1}}({\bm{x}})={{({\bm{w}}_k + {\bm{w}}_{k + 1})}^\top}{\bm{x}},
 \end{equation}
 where $k = 1, \ldots ,p - 1$, and proposed the following \emph{new hyperplane ranker},
\begin{equation}\label{NewPredictionFun}
r_{\rm{new}}: r({\bm{x}}) = 1 + \sum\limits_{i = 1}^{p - 1}I( {g_k}({\bm{x}}) > 0)
\end{equation}
where $I(\cdot)$ is the indicator function. One of the advantages of this prediction function (\ref{NewPredictionFun}) is that it can reduce the ambiguous region. Since the prediction is based on the ordered binary decomposition, the ordered structure information among labels is considered and can make the predicted label closer to the true label.

\begin{figure}[H]
  \centering
    \subfigure[Predict function $r_{\rm{old}}$]{
    \label{fig:demo:old} 
   \includegraphics[width=2in]{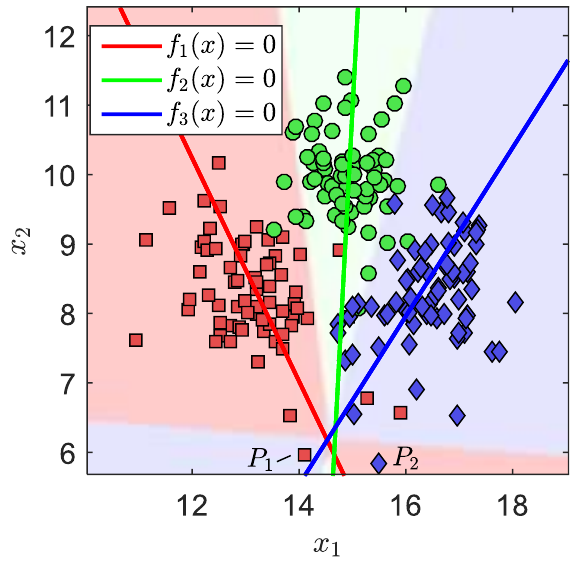}}
  \hspace{0.02in}
  \subfigure[Predict function $r_{\rm{new}}$]{
    \label{fig:demoe:rew} 
   \includegraphics[width=2in]{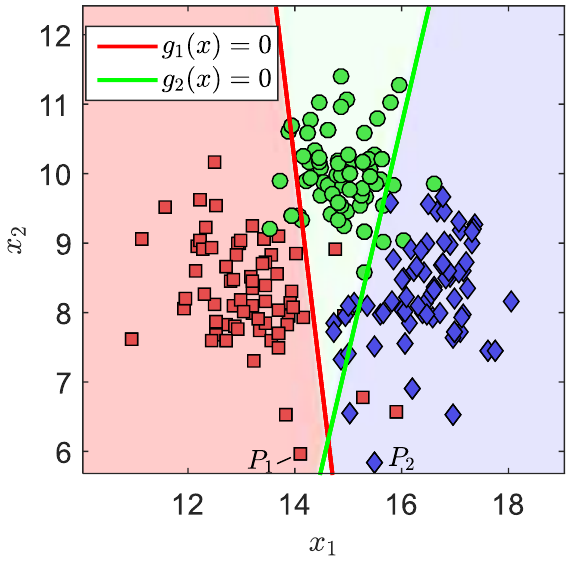}}
  \caption{Demo shows the difference of NPSVOR's two predict functions $r_{\rm{new}}$ and $r_{\rm{old}}$. The  points with square, circle and diamond shape stand for rank 1, rank 2 and rank 3, respectively. The decision regions is filled with different colors (rank 1(red),rank 2(green) and rank 3(blue)).}
\label{fig:prediction}
\end{figure}
%
See Figure \ref{fig:prediction}, the old predict function (\ref{decisionfun}) obviously produces a large ambiguous region, but it is eliminated by the new proposed predict function (\ref{NewPredictionFun}). Furthermore, two points $P_1$ and $P_2$ belonged to class 1 and class 3 are predicted incorrectly to rank 3 and rank 1 by the old predict function, which bring a large relative error. However, they are predicted rightly by the new prediction method.

\section{Numerical Experiments}\label{experiment}
In this section, we evaluate the proposed algorithm described in Sections \ref{NPSVOR} and compare it with other state-of-the-art SVM-based  ordinal regression models. Numerical experiments are carried out to test the performance of NPSVOR. All the implementations are in C++ based on LIBLINEAR and the experiments are conducted on a 64-bit machine with Intel Xeon 2.0GHz CPU (E5504), 4MB cache, and 8GB main memory. The programs used for our experiment is available at  \url{https://github.com/huadong2014/LinearNPSVOR/}. Two evaluation criteria are utilized to evaluate the prediction error between the predicted ordinal scales $\{ {\hat y_1}, \ldots ,{\hat y_n}\} $  and the true targets  $\{ {y_1}, \ldots ,{y_n}\} $: Mean Absolute Error (MAE), which is computed as
\begin{equation}\label{MAE}
{\rm{MAE}} = \frac{1}{n}\sum\limits_{i = 1}^n|\hat{y_i} - {y_i}|.
\end{equation}
and Mean Squared Error (MSE), which is computed as
\begin{equation}\label{MSE}
{\rm{MSE}} = \frac{1}{n}\sum\limits_{i = 1}^n (\hat{y_i} - {y_i})^2.
\end{equation}
which is widely used to evaluate the divergence between predicted sentiment labels and the ground truth labels \cite{Tang2015Document,Diao2014Jointly}.
\subsection{Datasets and experimental settings}
We consider the following datasets in our experiments:
\begin{itemize}
  \item  \textbf{TripAdvisor}\footnote{The dataset is available at \url{http://www.cs.virginia.edu/~hw5x/dataset.html}}, is a dataset about hotel reviews originally used in \cite{Wang2010Latent}. Each review is scored on a scale ranging from one star to five stars.
  \item  \textbf{YelpReviews} \cite{Zhang2016Character}, is obtained from the Yelp Dataset Challenge in 2015\footnote{\url{https://www.yelp.com/dataset_challenge}}.  One task of this Challenge is to predict the full number of stars the user has given.
  \item \textbf{Treebank}\footnote{\url{http://nlp.stanford.edu/sentiment/}}: Stanford Sentiment Treebank, is a classification task consisting of sentences extracted from movie reviews with human annotations of their sentiment. Train/dev/test splits are provided and each sentence is rated with a fine-grained label among five sentiment-level scores(\emph{very negative}, \emph{negative}, \emph{neutral}, \emph{positive}, \emph{very positive}).
  \item \textbf{MovieReview} \cite{Pang2005Seeing}, a collection of movie reviews whose numerical rating converted from a four-star system.\footnote{scale dataset v1.0: \url{http://www.cs.cornell.edu/people/pabo/movie-review-data/}}.
   \item \textbf{LargeMovie}\footnote{\url{http://ai.stanford.edu/~amaas/data/sentiment/}}, a movie review dataset for eight-class sentiment classification which contains substantially more data than the previous benchmark datasets. It provides a set of 25,000 highly polar movie reviews for training, and 25,000 reviews for testing. We combine the training dataset and testing dataset together in our experiment.
   \item \textbf{DoctorQuality},  a Chinese E-health dataset, we collected it from a doctor consulting APP which provides healthcare Q\&A service under the mobile platform. Patients can give the description about their diseases and this App will assign a doctor to answer and chat with the patients. This process proceeds in a text dialogue. Each patient can assess the doctor after their consultation. The assessment is given at five levels: \emph{Strong bad, Very dad, bad, satisfied and Very satisfied}.  In order to avoid the influence of extreme imbalance data, about 20k instances were extracted for each category. Jieba\footnote{Jieba is an open-source Chinese word segmentation tool which is available at \url{https://github.com/fxsjy/jieba} } were used for word segmentation.
   \item \textbf{Eight Amazon product datasets}, collected from two data resources: four datasets (\textbf{AmazonMp3}, \textbf{Video}, \textbf{Mobilephone}, \textbf{Camera}\footnote{
\url{http://sifaka.cs.uiuc.edu/~wang296/Data/index.html}}) are originally used in \cite{Wang2010Latent}, and another four datasets ((\textbf{Electronics}, \textbf{HealthCare}, \textbf{AppsAndroid}, \textbf{HomeKitchen}))\cite{mcauley2015image,mcauley2015inferring} are downloaded from Amazon product data\footnote{Amazon product reviews datasets: \url{http://jmcauley.ucsd.edu/data/amazon/}}.  Each review consists of a review raw text with a overall rating score(1-5 stars). We choose the different types of product review datasets  with different scales for our experiments. Since their rating distributions are very imbalanced, we down sample a balanced dataset from each category.
\end{itemize}

For all these English datasets, the following procedures are executed when preprocessing: stemming, stop word removal and omitting the words that occur less than three times, the term document frequency is larger than 50\% or is shorter than 2 in length. TF-IDF is used to extract text features from unigrams and bigrams. To examine the testing performance, we use a stratified selection to split each data set to 70\% training and 30\% testing. The number of instances, features, nonzero elements in training data, and our proposed measures tested on twenty-one datasets are reported in Table \ref{DataStatistics0}. To estimate the test performances MAE and MSE, a stratified selection is taken to split each data set into one fifth for testing and the rest for training.

\begin{table}[!htb]
\renewcommand{\arraystretch}{1.3}
\setlength{\tabcolsep}{1pt}
\caption{ Data statistics. $n$ and $m$ denote the numbers of instances and features in a
data set, respectively. Nonzeros indicates the number of non-zero entries.}
\label{DataStatistics0}
\centering
\small
\begin{tabular}[!htb]{lccccc}
\hline\noalign{\smallskip}
Datasets & Instances($n$) & Features($m$) & Nonzeros &Classes & Distribution\\
\hline
AmazonMp3&10391&65487&1004435&5&$\approx$2078\\
Video&22281&119793&1754092&5&$\approx$4456\\
Tablets&35166&201061&3095663&5&$\approx$7033\\
Mobilephone&69891&265432&5041894&5&$\approx$13978\\
Cameras&138011&654268&14308676&5&$\approx$27602\\
TripAdvisor&65326&404778&8687561&5&$\approx$13065\\
DoctorQuality&119879&325448&7265259&5&$\approx$23975\\
Treebank&11856&8569&98883&5&$\approx$2371\\
MovieReview&5007&55020&961379&4&$\approx$1251\\
YelpReview&1121671&3138663&102232013&5&$\approx$224334\\
LargeMovie&50000&309362&6588192&8&$\approx$6250\\
Electronics&409041&1422181&37303259&5&$\approx$81808\\
HealthCare&82251&283542&5201794&5&$\approx$16450\\
AppsAndroid&220566&253932&6602522&5&$\approx$44113\\
HomeKitchen&120856&427558&8473465&5&$\approx$24171\\
\noalign{\smallskip}\hline
\end{tabular}
\label{table_dataset}
\end{table}

\subsection{The Performance of NPSVOR}	
The parameter $C$ in NPSVOR is chosen in the range $\{2^{-5},2^{-4},\ldots,2^{5}\}$ by five-fold cross validation (CV) on the training set. The MAE was selected for parameter tuning.
After obtaining the optimal parameter $C$ of each dataset, we conduct training set with this $C$ and then obtain MAE/MSE on corresponding test sets.

The trained model by using the best parameter $C$ of each data set running on the whole training set under the best $C$ is then applied to obtain MAE/MSE on its corresponding test set. $\varepsilon$ in NPSVOR is fixed at 0.1. Since the results for all datasets are similar, we only show the results of the top eight datasets in Table \ref{table_dataset} in this section.				
			
\subsubsection{The Performance of NPSVOR on Two Prediction Functions}
NPSVOR is viewed as an extension of TSVM for ordinal regression. \cite{Wang2016Nonparallel} defined the prediction criterion as shown in (\ref{decisionfun}) which is based on the distance from the sample to the corresponding hyperplanes, i.e. the rank of nearest category is assigned to the label. However, this prediction criterion is originally proposed for standard (multi-)classification problems. But for OR problems, the order information contained in labels has not been considered. Inspired by \cite{lin2012reduction}, we propose a new prediction criterion as (\ref{NewPredictionFun}). The $p$ hyperplanes are learned for each class, and the label of a new instance is predicted by an ensemble of these ordered binary predictors. In order to verify the effectiveness of the new prediction function, we compare NPSVOR under the two prediction criteria on the above datasets. The experimental results are shown in Table \ref{CompareTwoRankers}.

\begin{table}[!htb]
\renewcommand{\arraystretch}{1.3}
\caption{Comparison of NPSVOR with the new prediction function ($r_{\rm{new}}$) and the old prediction function ($r_{\rm{old}}$). Best results are boldfaced. }
\label{CompareTwoRankers}
\centering
\small
\begin{tabular}{lccccc}
\hline\noalign{\smallskip}
 \multirow{2}{*}{Datasets} & \multicolumn{2}{c}{MAE} &&  \multicolumn{2}{c}{MSE}\\
 \cline{2-3}\cline{5-6}
 &$r_{\rm{old}}$&$r_{\rm{new}}$&&$r_{\rm{old}}$&$r_{\rm{new}}$\\
\hline
AmazonMp3&0.511&\textbf{0.508}&&0.720&\textbf{0.707}\\
Video&0.401&\textbf{0.390}&&0.543&\textbf{0.541}\\
Tablets&0.302&\textbf{0.299}&&0.412&\textbf{0.406}\\
Mobilephone&0.410&\textbf{0.404}&&\textbf{0.548}&0.554\\
Cameras&0.249&\textbf{0.247}&&0.328&\textbf{0.322}\\
TripAdvisor&0.366&\textbf{0.364}&&0.452&\textbf{0.447}\\
DoctorQuality&0.731&\textbf{0.727}&&1.304&\textbf{1.292}\\
Treebank&0.830&\textbf{0.827}&&1.242&\textbf{1.230}\\
MovieReview&\textbf{0.424}&0.424&&\textbf{0.452}&0.452\\
YelpReview&0.382&\textbf{0.379}&&0.475&\textbf{0.466}\\
LargeMovie&1.046&\textbf{1.022}&&2.274&\textbf{2.163}\\
Electronics&0.551&\textbf{0.548}&&0.775&\textbf{0.767}\\
HealthCare&0.617&\textbf{0.612}&&0.954&\textbf{0.941}\\
AppsAndroid&0.598&\textbf{0.594}&&0.879&\textbf{0.868}\\
HomeKitchen&0.543&\textbf{0.538}&&\textbf{0.766}&0.780\\
\noalign{\smallskip}\hline
\end{tabular}
\end{table}

From the Table \ref{CompareTwoRankers},  we can see that the performance of linear NPSVOR under the new predictor (\ref{NewPredictionFun}) is obviously better than that using old one (\ref{decisionfun}) on MAE and MSE.

\subsubsection{Effects of Different Parameters $C_1$  and $C_2$}

In previous experiments, we set the parameters $C_1$ and $C_2$ to an equal value for the NPSVOR. Now, we try to evaluate the performance when parameters $C_1$ and $C_2$ are set to different values. We still carry out our experiments on the above datasets.  $\varepsilon$ is fixed at 0.1. Grid search for $C_1$ and $C_2$ are tuned in the range ${\log _2}{C_1},{\log _2}{C_2}\in [ - 8, - 7, \ldots ,5]$ and 5-fold cross validation are conducted. The results of MSE and MAE are given in Table \ref{TableMAE}. Compared to the results obtained from setting $C_1$ and $C_2$ equally, tuning these two parameters respectively makes some improvement, but this improvement is not obvious. Therefore, $C_1=C_2$ is suggested for NPSVOR, so that it can reduce the time of tuning model parameters.

\subsubsection{Comparison of Two DCD Algorithms}
\begin{figure*}[!htb]
  \center
  \includegraphics[width=4.8in]{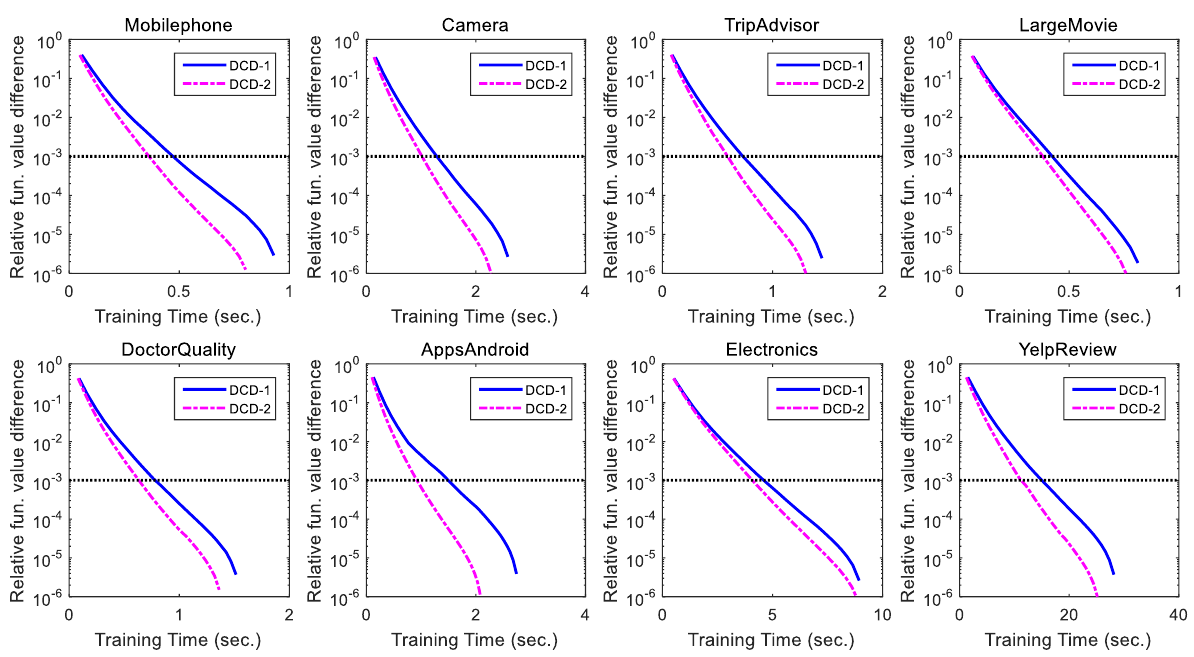}\\
  \caption{A comparison between two DCD algorithms (DCD-1 and DCD-2). We only present the training time and relative difference to the dual optimal function values for rank 3. NPSVOR with $C = 1$ and $ \varepsilon= 0.1$ is used. Both $x$-axis and $y$-axis are in log scale.}\label{fig:RelativeDifference}
\end{figure*}

Our first experiment is to compare two DCD implementations (Algorithms \ref{model:prim} and \ref{alg:LinearNPSVOR_detial}, called DCD-1 and DCD-2) so that only the better one is used for subsequent analysis. For this comparison, we consider NPSVOR with ${C_1} = {C_2} = 1$  and $\varepsilon  = 0.1$. Because the results for all data sets are similar, we only present the results for rank 3 of top four datasets in Figure \ref{fig:RelativeDifference}. The $x$-axis is the training time, and the $y$-axis is the relative difference to the dual optimal function value.
\begin{equation}\label{RelativeDiff}
\frac{{f({{\bm{\alpha }}_k}) - f({{\bm{\alpha }}^*})}}{{|f({{\bm{\alpha }}^*})|}}.
\end{equation}
where ${{\bm{\alpha }}^{\rm{*}}}$ is the optimum solution. We run optimization algorithms long enough to get an approximate $f({{\bm{\alpha }}^*})$.
Results show that DCD-2 is significantly faster than DCD-1. We notice that the running time in Figure \ref{fig:RelativeDifference} is log-scaled. This observation is consistent with our discussion in Section \ref{algorithm} that using DCD-1 directly suffers from some redundant operations.

\subsubsection{Role of The Parameter $\varepsilon$}

\begin{figure*}[h]
  \center
  \includegraphics[width=5.2in]{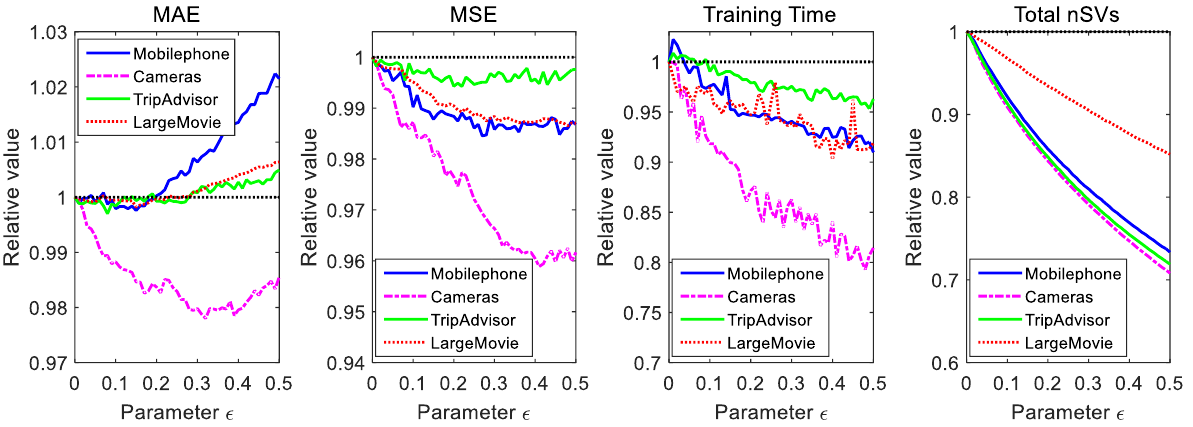}\\
  \caption{A comparison between NPSVOR with/without using $\varepsilon$. Show the MAE, MSE, Training time and the total number of support vectors(nSVs) of the model with $\varepsilon$ relative to it fix $\varepsilon=0$ on four datasets (Mobilephone, Cameras, TripAdvisor and LargeMovie)}\label{fig:Epsilon}
\end{figure*}

If  $\varepsilon {\rm{ = }}0$, linear NPSVOR can be reduced to
\begin{equation}\label{NoEpsilon}
\frac{1}{2}\left\| {\bm{w}}_k \right\|_2^2 + {C_1}\sum\limits_{i \in \mathcal{I}_k} {|{{\bm{w}}_k^\top}{{\bm{x}}_i}|}  + {C_2}\sum\limits_{i \notin \mathcal{I}_k} {\max \{ 0,1 - {{\tilde y}_{ki}}{{\bm{w}}_k^\top}{{\bm{x}}_i}\} }.
\end{equation}
The model becomes more concise. We are interested in the necessity of $\varepsilon$-insensitive loss function. DCD implementations can be applied to the situation of $\varepsilon =0$, so we conduct a comparison using DCD in Figure \ref{fig:Epsilon}. In order to verify the necessity of the parameter $\varepsilon$, we show the following four measures: MAE, MSE, cross validation time and the number of support vectors (SVs) with respect to $\varepsilon$ in $[0,0.5]$. We take four datasets (Mobilephone, Cammeras, TripAdvisor and Large Movie) to implement. During the experiment, 5-fold cross validation and grid search are used to get the optimal values of two penalty parameters  $C_1, C_2$  from $[{2^{ - 5}}, \ldots ,{2^5}]$ (set equal in experiments). We take $\varepsilon=0$ as the baseline and just show their changes of the relative values of dual objective function:
\[{\rm{MAE}}(\varepsilon )/{\rm{MAE}}(\varepsilon  = 0), {\rm{MSE}}(\varepsilon )/{\rm{MSE}}(\varepsilon  = 0),\]
\[ {\rm{Time}}(\varepsilon )/{\rm{Time}}(\varepsilon  = 0), {\rm{nSVs}}(\varepsilon )/{\rm{nSVs}}(\varepsilon  = 0)\]
As a reference, we draw a horizontal dotted line to indicate the relative difference $10^{-3}$.
For the comparison between NPSVOR with and without $\varepsilon$, Figure \ref{fig:Epsilon} indicates that NPSVOR with $\varepsilon$ can give a better MSE, faster running time and a more sparse solution. In fact, from the construction of NPSVOR, the  $\varepsilon$ insensitive loss helps to bring sparsity to the model's solution. The efficiency of DCD algorithm actually depends on the sparsity of the solution, the larger the value of $\varepsilon$ is, the sparser the solution is, and it runs faster. From Figure \ref{fig:Epsilon}, the relative values of running time on four datasets are decreasing with respect to $\varepsilon$ changing from 0 to 0.5. Considering the performance of MAE on the four datasets, three of them decrease first and then increase, getting the minimum value near $\varepsilon = 0.1$. But for the LargeMovie, the $\varepsilon$ in a proper interval can improve the value of MAE significantly, and the optimal value shows up at 0.2. Therefore, for these data sets, $\varepsilon$-insensitive loss function is useful and with $\varepsilon$, time for parameter selection can be reduced.

\subsection{Comparison of NPSVOR with Other Methods}

\cite{perez2014organ} is an outstanding review which contains comparisons among the existing state-of-the-art ordinal regression methods, like naive approach, threshold approach and ordered decomposition. The experiments showed that the SVM-based methods had a promising and better performance. In this experiment, four state-of-the-art SVM-based methods are selected for comparison.

We compare the generalization performance of the linear NPSVOR with the following four state-of-art SVM-based methods:
\begin{enumerate}
\item \textbf{SVC} \cite{Hsieh2008A}, a naive approach which treats the OR problem as a general multi classification problem. \cite{Hsieh2008A} provides a DCD algorithm for solving linear SVC, and takes two important accelerated strategies, the shrinking and random permutation. The algorithm can be founded in LIBLINEAR software package, using one vs. all strategy to deal with multi-classification problems.
\item \textbf{SVR} \cite{ho2012large}: The considered class of support vector regression was provided by the software implementation LIBLINEAR 3.21. The ordinal targets were treated as continuous values in standard SVR, and the predictions for test cases were rounded to the nearest ordinal scale. The insensitive zone parameter $\varepsilon$ of SVR was fixed at 0.1.
\item \textbf{SVOR} (Support Vector Ordinal Regression with Implicit Constraints of the ordered thresholds \cite{chu2007support}), which targets on finding the fixed max margin between each two nested classes. This method belongs to the threshold approach, where a real value mapping and several ordered thresholds were learned. Until now, there are still no large-scale algorithm for this method and just exists a nonlinear version.
\item \textbf{RedSVM} (Reduced to SVM) \cite{lin2012reduction}, a threshold approach to finding $(p-1)$ thresholds that divide the real-valued line into $p$ consecutive intervals which corresponds to the $p$ ranks. \cite{lin2012reduction} proposed a reduction framework which reduces the ordinal regression to binary classification.
\end{enumerate}

In order to analyze the performance of these methods in different scale datasets and their differences between the linear and nonlinear cases in dealing with high dimensional sparse problems, two experiments are carried out as following:

\subsubsection{Comparison of Linear/Nonlinear Models on Small Datasets}
 Some models still have no implementation for large-scale cases, such as SVOR \cite{chu2007support} and RedSVM \cite{lin2012reduction}, which only have their nonlinear versions (kernelized dual models solved by SMO),  we need to check whether those models with linear cases could give a competitive performance compared to their nonlinear cases for some applications, and could also enjoy faster training.

 For all the nonlinear models, we consider RBF kernel $\kappa(\bm{x}_1,\bm{x}_2)=\exp(-\gamma\|\bm{x}_1-\bm{x}_2\|^2)$, where $\gamma$ is a user-specified parameter. Because the training of nonlinear SVM-based models is time consuming, we only use a subset of each dataset for training.  The number of each subset is sampled less than 10000 from each dataset with a certain proportion as shown in the first column of Table \ref{Tables:smalldata}.  All subsets are divided into two parts, the 70\% of data for training and the rest for testing.  We conduct five-fold cross validation (CV) to find the best $C\in \{2^{-5},\ldots,2^5\}$ and $\gamma \in \{2^{-8},2^{-7},\ldots,2^0\}$. We denote the NPSVOR model with linear kernel as Lin-NPSVOR and denote it with the nonlinear kernel (RBF) as NPSVOR. The MAE/MSE are reported in Table \ref{Tables:smalldata}.  The average ranking of each method on MAE/MSE are also presented in the last row of the table.

\begin{table*}[!htb]
\caption{Test MAE and MSE results on the testing set for each small dataset and methods, including the results of these five nonlinear svm-based models. Boldface the best results.}
\setlength{\tabcolsep}{1.8pt}
\label{Tables:smalldata}
\centering
\footnotesize
\begin{tabular}{lcccccccccccccc}
\hline\noalign{\smallskip}
 \multirow{2}{*}{Datasets} & \multicolumn{6}{c}{MAE} &&  \multicolumn{6}{c}{MSE}\\
 \cline{2-7}\cline{9-14}
&SVC&SVR&SVOR&{\scriptsize REDSVM}&{\scriptsize NPSVOR}& {\scriptsize Lin-NPSVOR}&&SVC&SVR&SVOR&{\scriptsize REDSVM}&{\scriptsize NPSVOR}& {\scriptsize Lin-NPSVOR}\\
\hline
AmazonMp3(10\%)&0.803&0.800&0.761&0.761&\textbf{0.745}&0.748&&1.501&1.125&1.127&1.127&\textbf{1.073}&1.083\\
Video(10\%)&0.679&0.737&0.674&0.674&0.651&\textbf{0.647}&&1.115&0.993&0.896&0.895&0.875&\textbf{0.872}\\
Tablets(10\%)&0.582&0.666&0.580&0.579&\textbf{0.563}&0.564&&0.916&0.833&0.746&0.743&\textbf{0.727}&0.727\\
Mobilephone(10\%)&0.620&0.697&0.624&0.623&0.607&\textbf{0.607}&&1.009&0.919&0.827&0.829&\textbf{0.809}&0.812\\
Cameras(1\%)&0.752&0.815&\textbf{0.728}&0.729&0.738&0.739&&1.251&1.094&\textbf{0.955}&0.956&0.982&0.988\\
TripAdvisor(10\%)&\textbf{0.515}&0.620&0.563&0.563&0.518&0.521&&0.757&0.755&0.698&0.699&0.640&\textbf{0.639}\\
DoctorQuality(10\%)&1.240&1.086&1.082&1.080&\textbf{1.059}&1.061&&2.701&1.780&1.779&1.777&\textbf{1.765}&1.774\\
Treebank&1.009&0.933&\textbf{0.924}&0.976&0.928&0.930&&1.858&1.421&\textbf{1.380}&1.740&1.441&1.462\\
MovieReview&0.616&0.569&0.560&\textbf{0.558}&0.562&0.560&&0.758&0.627&0.618&\textbf{0.616}&0.629&0.627\\
YelpReview(0.1\%)&\textbf{0.747}&0.946&0.892&0.870&0.791&0.792&&1.381&1.404&1.290&1.250&\textbf{1.063}&1.069\\
LargeMovie(10\%)&1.360&1.374&1.286&1.286&\textbf{1.198}&1.198&&4.462&3.051&2.752&2.752&2.618&\textbf{2.616}\\
Electronics(1\%)&0.899&0.892&0.840&0.841&\textbf{0.816}&0.819&&1.779&1.305&1.256&1.257&\textbf{1.187}&1.198\\
HealthCare(10\%)&0.866&0.866&0.820&0.820&\textbf{0.801}&0.801&&1.735&1.274&1.285&1.285&\textbf{1.233}&1.260\\
AppsAndroid(1\%)&0.909&0.891&\textbf{0.824}&0.824&0.832&0.835&&1.758&1.286&1.224&1.224&\textbf{1.212}&1.224\\
HomeKitchen(1\%)&0.901&1.022&0.970&0.971&0.914&\textbf{0.900}&&1.736&1.557&1.465&1.466&1.342&\textbf{1.328}\\
\hline
Avg. Ranking&4.600&5.467&3.200&3.267&\textbf{1.933}&2.267&&5.933&4.467&3.067&3.267&\textbf{1.800}&2.067\\
\noalign{\smallskip}\hline
\end{tabular}
\end{table*}

\subsubsection{Comparison of Linear Methods on Large Datasets}\label{subsect:Linear Methods}
Since no large-scale algorithm has been proposed for linear RedSVM, we modify the DCD algorithm which is originally designed for linear SVM to solve this model (RedSVM can be reduced to a binary classification based on SVC by extending the samples \cite{lin2012reduction}). The parameter $C$ is chosen in range $\{2^{-5},2^{-4},\ldots,2^{5}\}$ by five-fold cross validation (CV) on the training set. Using models trained under the best $C$, we conduct prediction on the testing set to obtain the testing MAE and MSE. We use stopping tolerance 0.1 for SVC, RedSVM and NPSVOR methods although their stopping conditions are slightly different, and use the default stopping tolerance 0.01 for SVR in LIBLINEAR. $\varepsilon$ in NPSVOR and SVR is fixed to the value of 0.1.  MAE was used for tuning the parameter(s).

Our code supports a bias term in a common way by appending one more feature to each data instance. We apply it to all the data sets, just compare the MAE and MSE values with the bias term. With the stopping tolerance $\varepsilon = 0.1$, the results in Table \ref{TableMAE} show that the values obtained with bias term are similar for almost all the data sets. The average ranks of each methods on MAE and MSE are presented in the last row of Table \ref{TableMAE}.  We also provide the training time of those models in Table \ref{TrainingTime}  which is running on training set under the best parameter $C$ tuned on each dataset.

\begin{table}[!htb]
\renewcommand{\arraystretch}{1.3}
\caption{The MAE/MSE on 15 datasets of four methods in their linear cases. Boldface the best results.}
\setlength{\tabcolsep}{1pt}
\label{TableMAE}
\centering
\small
\begin{tabular}{lccccccccc}
\hline\noalign{\smallskip}
 \multirow{2}{*}{Datasets} & \multicolumn{4}{c}{MAE} &&  \multicolumn{4}{c}{MSE}\\
 \cline{2-5}\cline{7-10}
&SVC&SVR&RedSVM&NPSVOR&&SVC&SVR&RedSVM&NPSVOR\\
\hline
AmazonMp3&0.595&0.561&0.556&\textbf{0.508}&&1.095&0.757&0.755&\textbf{0.707}\\
Video&0.436&0.445&0.464&\textbf{0.390}&&0.756&0.591&0.620&\textbf{0.541}\\
Tablets&0.325&0.344&0.363&\textbf{0.299}&&0.551&0.447&0.472&\textbf{0.406}\\
Mobilephone&0.442&0.463&0.464&\textbf{0.404}&&0.746&0.607&0.608&\textbf{0.554}\\
Cameras&0.269&0.296&0.290&\textbf{0.247}&&0.435&0.380&0.378&\textbf{0.322}\\
TripAdvisor&0.405&0.443&0.429&\textbf{0.364}&&0.625&0.533&0.521&\textbf{0.447}\\
DoctorQuality&0.793&0.860&0.803&\textbf{0.727}&&1.763&1.348&1.374&\textbf{1.292}\\
Treebank&0.937&0.823&\textbf{0.803}&0.827&&1.701&1.175&\textbf{1.159}&1.230\\
MovieReview&0.474&0.449&0.429&\textbf{0.424}&&0.565&0.465&0.454&\textbf{0.452}\\
YelpReview&0.462&0.469&0.462&\textbf{0.379}&&0.559&0.580&0.559&\textbf{0.466}\\
LargeMovie&1.213&1.213&1.130&\textbf{1.022}&&3.651&2.531&2.368&\textbf{2.163}\\
Electronics&0.611&0.619&0.632&\textbf{0.548}&&1.112&0.860&0.866&\textbf{0.767}\\
HealthCare&0.645&0.680&0.695&\textbf{0.612}&&1.333&1.032&1.058&\textbf{0.941}\\
AppsAndroid&0.656&0.667&0.665&\textbf{0.594}&&1.220&0.929&0.936&\textbf{0.868}\\
HomeKitchen&0.613&0.599&0.633&\textbf{0.538}&&1.126&0.831&0.866&\textbf{0.780}\\
\hline
Avg. Ranking&2.533&3.267&3.000&\textbf{1.133}&&3.867&2.467&2.467&\textbf{1.133}\\
\noalign{\smallskip}\hline
\end{tabular}
\end{table}

\begin{table}[!htb]
\renewcommand{\arraystretch}{1.3}
\caption{Training time of five methods under the best parameter(s).}
\label{TrainingTime}
\centering
\small
\begin{tabular}{lcccc}
\hline\noalign{\smallskip}
Datasets&SVC&SVR&RedSVM&NPSVOR\\
\hline
AmazonMp3&0.167&0.043&0.205&0.128\\
Video&0.544&0.085&0.372&0.270\\
Tablets&1.307&0.162&0.664&0.496\\
Mobilephone&2.305&0.333&1.544&1.103\\
Cameras&11.854&1.193&5.348&3.352\\
TripAdvisor&4.488&0.511&1.892&1.257\\
DoctorQuality&3.135&0.289&1.612&2.286\\
Treebank&0.012&0.006&0.010&0.061\\
MovieReview&0.077&0.025&0.065&0.074\\
YelpReview&57.501&16.044&28.862&32.389\\
LargeMovie&5.815&0.283&0.952&1.362\\
Electronics&34.508&4.746&10.462&11.319\\
HealthCare&2.840&0.470&1.603&0.945\\
AppsAndroid&6.013&0.493&1.757&1.998\\
HomeKitchen&5.446&0.833&2.205&2.422\\
\noalign{\smallskip}\hline
\end{tabular}
\end{table}

From the Table \ref{TableMAE} and Table \ref{TrainingTime}, we can observe the following results:
\begin{enumerate}
\item The NPSVOR  almost achieves the best overall results compared to the SVC, SVR and RedSVM on the MAE and MSE. It shows that the prediction result of NPSVOR is closer to the true label and this method reduces the possibility of a larger prediction deviation.
\item The performance of linear RedSVM is inferior to the RedSVM of the non-linear case \cite{chu2007support}. However, from the experimental results,  its performance is rather commonplace and even inferior to the standard SVC on some data sets. The underlying assumption of threshold methods is that all classes are well ordered in a unique direction and separable by parallel hyperspaces in the primal feature space. This assumption could not be always satisfied without using nonlinear feature mapping.
\item Both NPSVOR and SVC train an independent optimization model for each category, but with the same termination accuracy, NPSVOR has a faster training speed than SVC. Because they use different shrinking strategies and termination conditions in their algorithm design.  In these methods, SVR achieves the fastest training speed because it only needs to solve an optimization model of the same size as the SVC subproblem.
\end{enumerate}

\begin{figure*}[!htb]
  \center
    \subfigure[Mobilephone]{
    \label{fig:demo:1} 
   \includegraphics[width=5.2in]{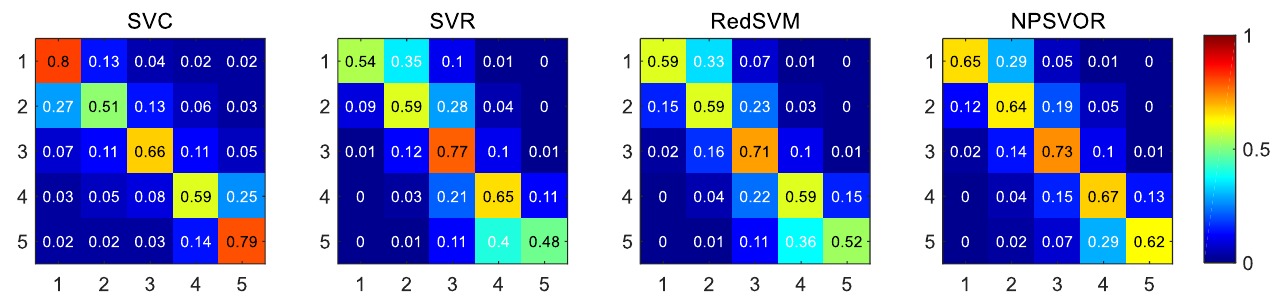}}
  \subfigure[Camera]{
    \label{fig:demoe:2} 
   \includegraphics[width=5.2in]{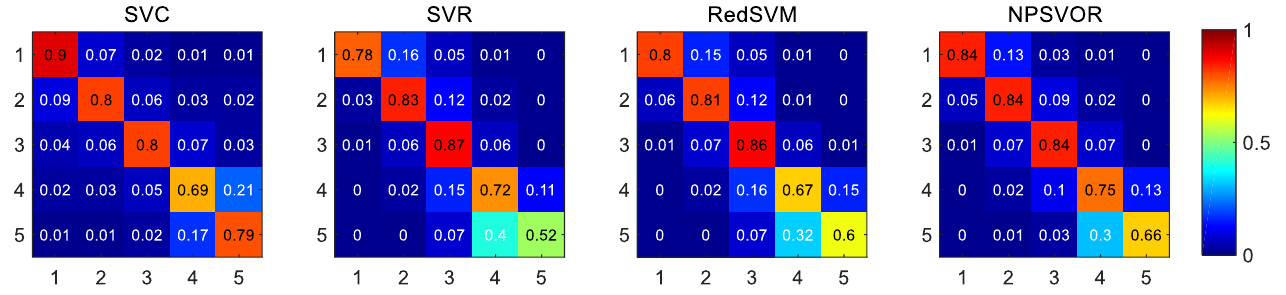}}
  \caption{Confusion matrices of four methods under two datasets (MobilePhone and Camera)}\label{fig:ConfusionMatrix}
\end{figure*}

Ordinal regression is different from the normal multiclass classification, it needs the predicted label as close as possible to the actual label. To show the difference between the prediction of the four models, we visualize the confusion matrix of the prediction results.
Figure \ref{fig:ConfusionMatrix} shows the confusion matrixes of the four methods on the datasets in Table \ref{DataStatistics0}, where the rows refer to the real ratings and the columns refer to the assigned ranks. We only show two datasets \emph{Mobilephone} and \emph{Camera} for
discussion here; other datasets exhibit a similar results.

For the dataset ``Mobilephone", SVC assigns the samples with rank 1 wrongly to rank 5 with an error rate of 2\%, while NPSVOR does not assign any samples of rank 1 to rank 5. Since the order information existing among labels is considered in modeling, the performance of SVR, RedSVM and NPSVOR are in line with expectations, i.e. the predicted label is closer to the true label than SVC. Compared with SVC, although both SVR and RedSVM improve the MAE/MSE with a certain extent, they get a lower prediction accuracy. NPSVOR achieves  a relatively higher prediction accuracy. From the diagonal of each matrix, the prediction accuracy of NPSVOR for each category is significantly higher than SVC on intermediate
ranks (rank 2,3,4), and SVC only gets higher accuracy at both end sides (rank 1 and rank 5).

\subsection{Sensitivity of Parameter $C$}
\begin{figure*}[!htb]
  \center
  \includegraphics[width=5.2in]{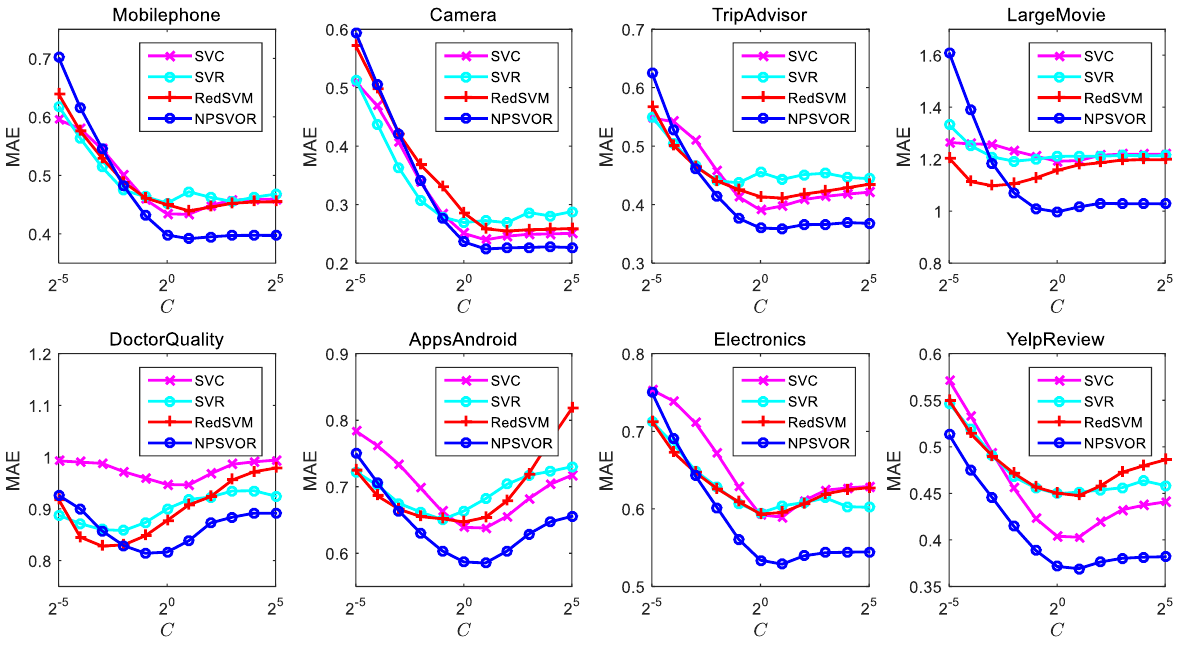}\\
  \caption{MAE changes with the parameter $C$ on eight datasets}\label{fig:MAEwithC}
\end{figure*}

To observe the sensitivity of the parameter $C$ in models, we compare the four methods (SVC, SVR, RedSVM and NPSVOR) on the prediction performance and the training time. The value of $C$ is changed in the range $\{ {2^{ - 5}},{2^{ - 4}}, \ldots ,{2^5}\} $. For each value $C$, we train the models on each training set and then apply the trained models to the testing set. The parameter settings are the same as Section \ref{subsect:Linear Methods}.  Figure \ref{fig:MAEwithC} show how the MAE change with respect to $C$, respectively. It can be observed that MAE of NPSVOR changes stably with $C$. When the value of $C$ is larger than 0.1, the performance of NPSVOR is obviously better than other three methods. Since the figure under MSE is similar with Figure \ref{fig:MAEwithC}, we don't discuss and show it again.

\section*{Conclusion and further works}\label{conclusion}

 We have studied the NPSVOR in linear case for training large-scale ordinal regression problems. A new DCD method (DCD-2) having the same efficiency with linear SVC is proposed to solve a reformulation of the dual problem of NPSVOR. It utilizes the structural relationship of the solution of the dual model and then reformulates the dual model with only $n$ variables. Empirical comparisons show that the training time of DCD-2 reduces obviously when compared to DCD-1.  The proposed method can train linear NPSVOR efficiently and achieve comparable performances to other state-of-the-art methods. To further utilize the ordered information contained in labels, a new predict function $r_{\rm{new}}$ is defined, which is obtained by $p-1$ ordered binary discrimination functions constructed by $p$ learned hyperplanes.  Experiments show that $r_{\rm{new}}$ is better than the OR predictor $r_{\rm{old}}$ based on the minimum distance principle.

 In addition, although we only consider the hinge loss in our implementation for the experiments because of the lack of space, our method is applicable to the square hinge loss (L2 loss), logistic loss etc with a minor change. Note that each hyperplane learned in NPSVOR is obtained independently with each other, without considering their correlation. Considering OR as a structured and interdependent output problem is our future work.

\subsection*{Acknowledgments}
This work is partially supported in part by the National Science Foundation of China under Grant No.11671379, No.11331012, No.71331005, No.91546201.

\bibliographystyle{unsrt}
\bibliography{mybibliography}

\end{document}